\documentclass[letterpaper, 10 pt, conference]{ieeeconf}  
\IEEEoverridecommandlockouts                              

\usepackage{graphics} 
\usepackage{epsfig} 
\usepackage{amsmath} 
\usepackage{color}
\usepackage{multirow}
\usepackage[breaklinks,colorlinks,linkcolor=black,citecolor=black,urlcolor=black]{hyperref}
\usepackage{diagbox}
\usepackage{cite}
\usepackage{caption}
\usepackage[ruled]{algorithm2e}
\usepackage{amsfonts,amssymb}
\usepackage{amssymb}
\usepackage{mathrsfs}
\usepackage{pifont}
\newcommand{\cmark}{\ding{51}}%
\newcommand{\xmark}{\ding{53}}%
\DeclareMathOperator*{\argmin}{argmin}
\begin{document}

\title{\LARGE \bf
	An Image-based Approach of Task-driven Driving Scene Categorization
}
\author{Shaochi~Hu$^1$, Hanwei~Fan$^1$, Biao~Gao$^1$, Xijun~Zhao$^2$
	and Huijing~Zhao$^1$, ~\IEEEmembership{Member,~IEEE}
	\thanks{This work is supported in part by the National Natural Science Foundation of China under Grant 61973004}
	\thanks{$^1$S.Hu, H.Fan, B.Gao and H.Zhao are with the Key Lab of Machine Perception (MOE), Peking University, Beijing, China. $^2$X.Zhao is with China North Vehicle Research Institute, Beijing, China}
	\thanks{Correspondence: H. Zhao, zhaohj@cis.pku.edu.cn.}
}

\maketitle
\thispagestyle{empty}
\pagestyle{empty}

\begin{abstract}
	Categorizing driving scenes via visual perception is a key technology for safe driving and the downstream tasks of autonomous vehicles. 
	Traditional methods infer scene category by detecting scene-related objects or using a classifier that is trained on large datasets of fine-labeled scene images.
	Whereas at cluttered dynamic scenes such as campus or park, human activities are not strongly confined by rules, and the functional attributes of places are not strongly correlated with objects. So how to define, model and infer scene categories is crucial to make the technique really helpful in assisting a robot to pass through the scene.
	This paper proposes a method of task-driven driving scene categorization using weakly supervised data.
	Given a front-view video of a driving scene, a set of anchor points is marked by following the decision making of a human driver, where an anchor point is not a semantic label but an indicator meaning the semantic attribute of the scene is different from that of the previous one.
	A measure is learned to discriminate the scenes of different semantic attributes via contrastive learning, and a driving scene profiling and categorization method is developed based on that measure.
	Experiments are conducted on a front-view video that is recorded when a vehicle passed through the cluttered dynamic campus of Peking University. The scenes are categorized into straight road, turn road and alerting traffic. The results of semantic scene similarity learning and driving scene categorization are extensively studied, and positive result of scene categorization is 97.17 \% on the learning video and 85.44\% on the video of new scenes.
	
\end{abstract}

\section{INTRODUCTION}

	Human activities are strongly connected with the semantic attributes of scene objects and regions. For example, the zebra-zone defines a region where pedestrians cross the road; traffic signal prompts an intersection where a mixture of traffic flows happen, etc.
	Categorizing driving scene via environmental perception is a key technology for safe driving and the downstream tasks of autonomous vehicles such as object detection, behavior prediction and risk analysis, etc.
	
	Place/scene categorization \cite{sunderhauf2016place} or scene classification \cite{rangel2016scene} belongs to the same group of researches, which has been studied broadly in the robotic field such as semantic mapping \cite{kostavelis2015semantic} or navigating a robot using scene-level knowledge \cite{kostavelis2016robot}.
	Many methods infer scene category by detecting certain objects that have unique functions in specific scenes, such as bed for a bedroom or knife for a kitchen \cite{liao2016understand}\cite{pronobis2012large}.
	Deep learning methods have been the vast majority of the researches in recent years, where a CNN is trained to map the input image into a scene label, and supervised learning \cite{narayanan2019dynamic} are conducted mostly which require a large amount of fine-labeled datasets.

\begin{figure}[t]
	\centering
	\includegraphics[keepaspectratio=true,width=\linewidth]{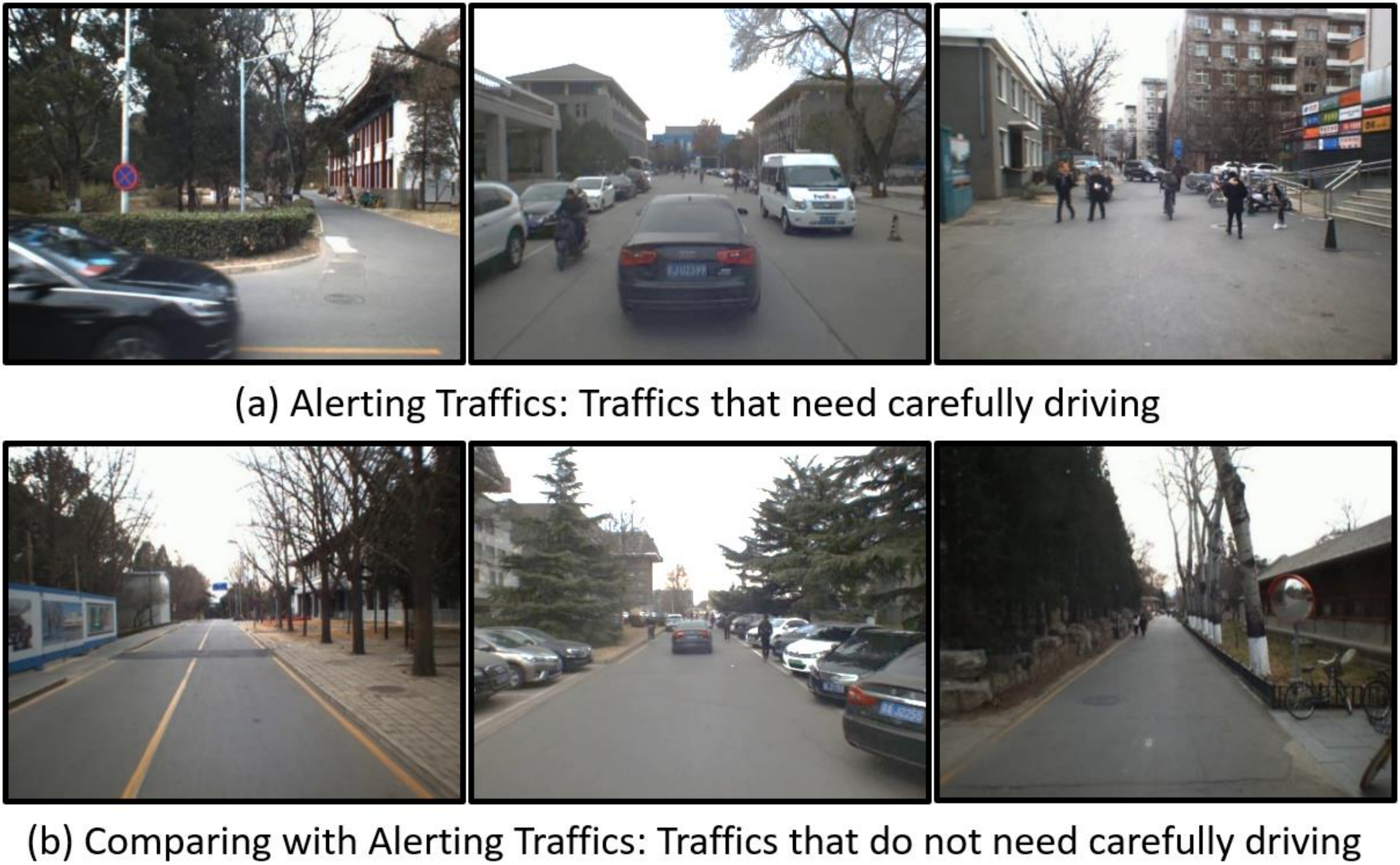}
	\caption{The difficulty of categories definition in cluttered dynamic scenes.}
	\vspace{-6mm}
	\label{fig:alertingTraffic}
\end{figure}
	
	MIT place\cite{zhou2014learning} and Large Scene understanding (LSUN)\cite{yu2015lsun} datasets contain a large number of labeled images that were originally downloaded from websites by using Google and other image search engines. Honda \cite{narayanan2019dynamic} and FM3\cite{sikiric2019traffic} datasets are more motivated by robotic and autonomous driving applications, which contain egocentric videos of driving scenes that are labeled at the frame level.
	Scene labels (i.e. categories) of these datasets are mainly defined on the following aspects: (1) functional attribute of places, e.g. cafeteria, train station platform \cite{zhou2014learning}, intersection, construction zone \cite{narayanan2019dynamic}; (2) a target object of the scene, e.g. people, bus \cite{yu2015lsun}, overhead bridge \cite{sikiric2019traffic}; (3) the static status of the target object or scene, e.g. road surface, weather condition \cite{narayanan2019dynamic}; (4) the dynamic status of the agent with respect to the scene, e.g. approaching, entering and leaving an intersection\cite{narayanan2019dynamic}.
	
	However, scene categorization based on the above category definitions may be less helpful for a robot to traverse cluttered dynamic urban scenes.
	For scenes shown in Fig. \ref{fig:alertingTraffic}, the functional attributes of places are not clearly defined, e.g. pedestrians can walk on sidewalk or motorway, intersections have neither traffic signal nor zebra-zone, and pedestrians cross the road anywhere. On the other hand, categorizing scenes on the presence of certain objects is less meaningful as the scenes are populated, complex and are usually not associated with any special objects. For example, pedestrians, cyclists or cars may appear anywhere, whereas the agent needs to make a special handle only when potential risks are detected.
	How to define semantically meaningful scene categories is crucial to make the scene categorization results really helpful in the tasks such as assisting human drivers or autonomous driving at cluttered dynamic urban scenes.
	
	This research proposes a task-driven driving scene categorization method by learning from the human driver's decision making. The normal state of an experienced human driver is usually relaxed. He/she is alerted when the situation is different from normal, such as a potential danger, or when the situation changes from the early moment (i.e. scenes with different semantic attributes), such as approaching an intersection. Given a video of driving scene, we let a human driver mark the scenes with sparse anchor points, where a pair of successive anchor points denote scenes of different semantic attributes. Inspired by the recent progress on contrastive learning \cite{chen2020big} and the impressive results in visual feature representation \cite{kolesnikov2019revisiting} and metrics learning \cite{lu2017deep}, our research learns a semantic scene similarity measure via anchor points using contrastive learning, which is exploited to profile scenes of different categories, and further infer scene labels for both online and offline applications such as semantic mapping and driving assistant. Experiments are conducted using a front-view video that was collected when a vehicle drived through a cluttered dynamic campus. The scenes are categorized into straight road (normal driving with no alerting traffic), turn road and alerting traffic. The proposed method is extensively examined by experiments.

\begin{figure*}[h]
	\centering
	\includegraphics[keepaspectratio=true,width=\linewidth]{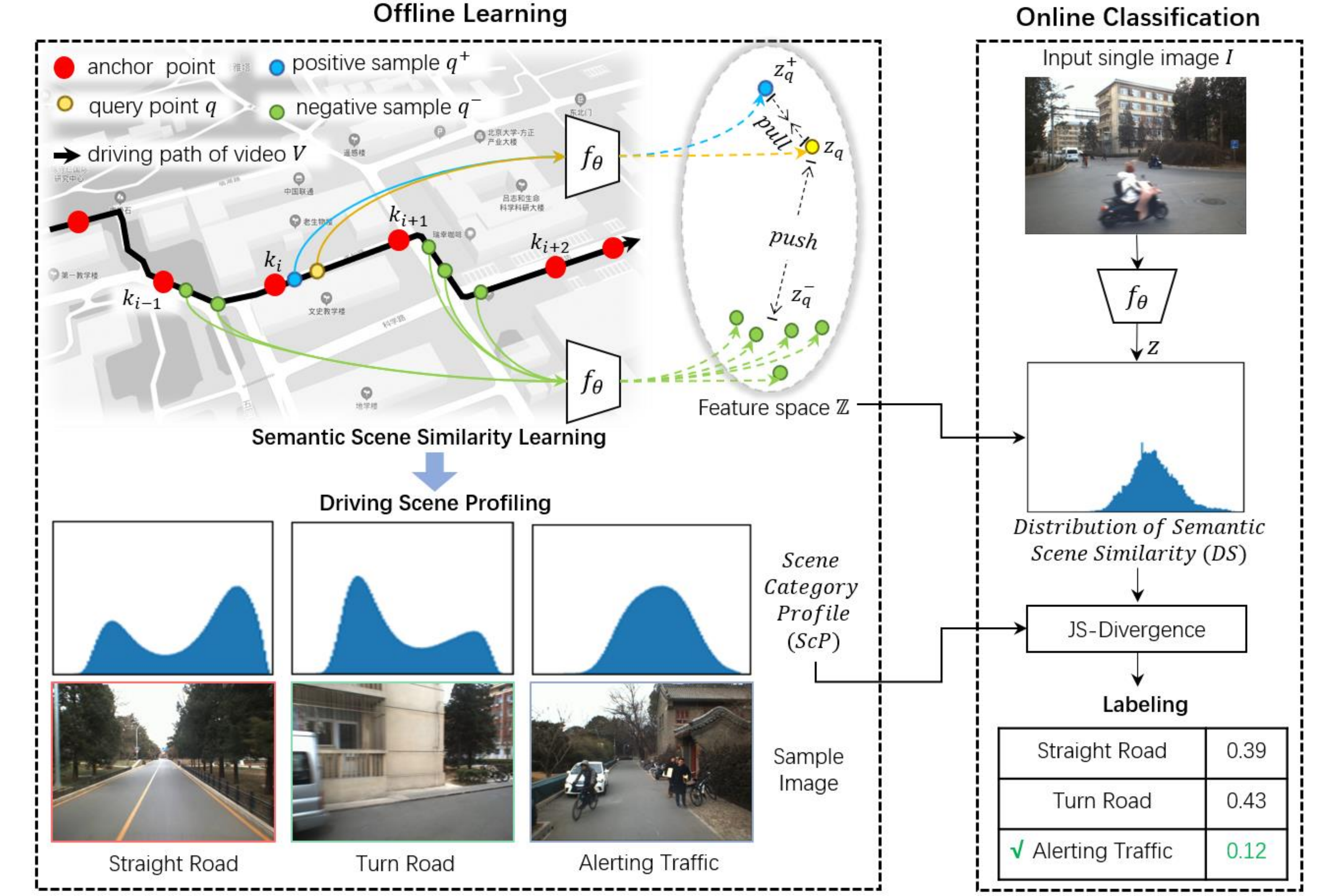}
	\caption{Algorithm pipeline. The training detail of $f_\theta$ is shown in Sec. \ref{section:sssl}, and the above $f_\theta$ share weights. The implementation of DS and S$c$P are shown in Sec. \ref{section:d3s} and \ref{section:scp}.}
	\vspace{-4mm}
	\label{fig:meth}
\end{figure*}	


\section{Related Works}\label{section:relatedwork}

\subsection{Scene Categorization Method}

Many methods infer scene category by detecting certain objects that have unique functions in specific scenes. \cite{pronobis2012large} combines object observations, the shape, size, and appearance of rooms into a chain-graph to represent the semantic information and perform inference. \cite{viswanathan2011place} learns object-place relations from an online annotated database and trains object detectors on some of the most frequently occurring objects by hand-crafted features. \cite{espinace2013indoor} uses common objects, such as doors or furniture, as a key intermediate representation to recognize indoor scenes.

Motivated by the success of deep learning in addressing visual image recognition tasks, CNN has been widely used as a feature encoder or mapping function of image and label for scene categorization in robotics. \cite{urvsivc2016part} presents a part-based model and learns category-specific discriminative parts for the part-based model. \cite{mancini2017learning} represents images as a set of regions by exploiting local deep representations for indoor image place categorization. \cite{zrira2018discriminative} trains a discriminative deep belief network (DDBN) classifier on the GIST features of the image.
Driving scene categorization is conducted in \cite{wu2017traffic} by combining the features
representational capabilities of CNN with the VLAD encoding scheme, and \cite{narayanan2019dynamic} trains an end-to-end CNN for this task. 

\subsection{Scene Category Definition}
	Here we review some typical scene categorization datasets, and we focus on their label definition, data acquisition and annotation method.
	
	Honda \cite{narayanan2019dynamic} and FM3\cite{sikiric2019traffic} datasets focus on driving scene categorization, and Honda defines 12 typical traffic scenarios: {\it branch left, branch right, construction zone, merge left, 3\&4\&4-way intersection, overhead bridge, rail crossing, tunnel, zebra crossing}, where FM3 defines 8 traffic scene: {\it highway, road, tunnel, exit, settlement, overpass, booth, traffic}. They are both collected by an on-board RGB camera and annotated by human operators. Their label definition focuses on the road static structure regardless of the dynamic of traffic. 
	
	MIT place\cite{zhou2014learning} and Large Scene understanding (LSUN)\cite{yu2015lsun} datasets contain a large number of labeled images that were originally downloaded from websites by using Google and other image search engines, and then annotated by human. MIT place dataset has up to 476 fine-grained scene categories including from indoor to outdoor such as {\it romantic bedroom, teenage bedroom, bridge and forest path}, etc. LSUN has only 10 scene categories: {\it bedroom, kitchen, living room, dining room, restaurant, conference room, bridge, tower, church, outdoor}. They are labeled mainly by the functional attribute of a place and the target object of the scene.
	
	The above categories definition may be less helpful for our task-driven driving scene categorization problem which focuses on cluttered dynamic urban scenes.
	
\subsection{Contrastive Learning}	

	Self-supervised learning\cite{jing2020self} aims at exploring the internal distribution of data by constructing a series of tasks using different prior knowledge of images. Recent progress in self-supervised learning mainly contributed by contrastive learning\cite{chen2020big}, which effectively represents image feature by training InfoNCE loss\cite{oord2018representation} on delicately constructed positive and negative samples. Memory-based method \cite{wu2018unsupervised} \cite{tian2019contrastive} suffer from the old feature in memory resulting in less consistent for the update of the network, and the performance of end-to-end approach\cite{ye2019unsupervised, hjelm2018learning, bachman2019learning} are limited by the GPU memory. Moco \cite{he2020momentum} makes a balance between feature updating and memory size by maintaining a queue and momentum updating the network. SimCLR\cite{chen2020simple} improves the performance by applying a projection head at the end of the network.
	
\section{Methodology}\label{section:methodology}

\subsection{Problem Formulation}\label{section:problem}

A front-view video $V$ of a driving scene is marked by a human driver with a set of anchor points $K=\{k_1,k_2,...,k_n\}$, where $k_i$ is a frame index of the video, and an anchor point $k_i$ means that the semantic attribute $A$ of the driving scene captured by image frame $V(k_i)$ is different from that of the previous anchor point $k_{i-1}$, i.e. $A(k_i) \neq A(k_{i-1})$.
Without loss of generality, we assume that the driving scenes captured by video $V$ is temporally continuous, i.e. $A(k_i) = A(k_{i}+\delta)$ and $A(k_i) \neq A(k_{i-1}+\delta)$, where $\delta$ is a small integer indicating the neighbor frames of an anchor point ({\it neighborhood consistent assumption}).

Given a video $V$ and a set of anchor points $K$ for training, this work is to learn a scene profiler to model the driving scenes of different semantic categories $\mathbb{C}$ such as straight road and turn road, and further infer scene labels for both online and offline applications such as semantic mapping and driving assistant.

To make the scene labels semantically meaningful, the categories of semantic scenes $\mathbb{C}$ are predefine by a human operator, which is represented by a typical image frame $q_c$ for each scene category $c$.

This research studies method of using a single image as input, which could be extended in the future by inferring on video clips.

\begin{figure*}[]
	\centering
	\includegraphics[keepaspectratio=true,width=\linewidth]{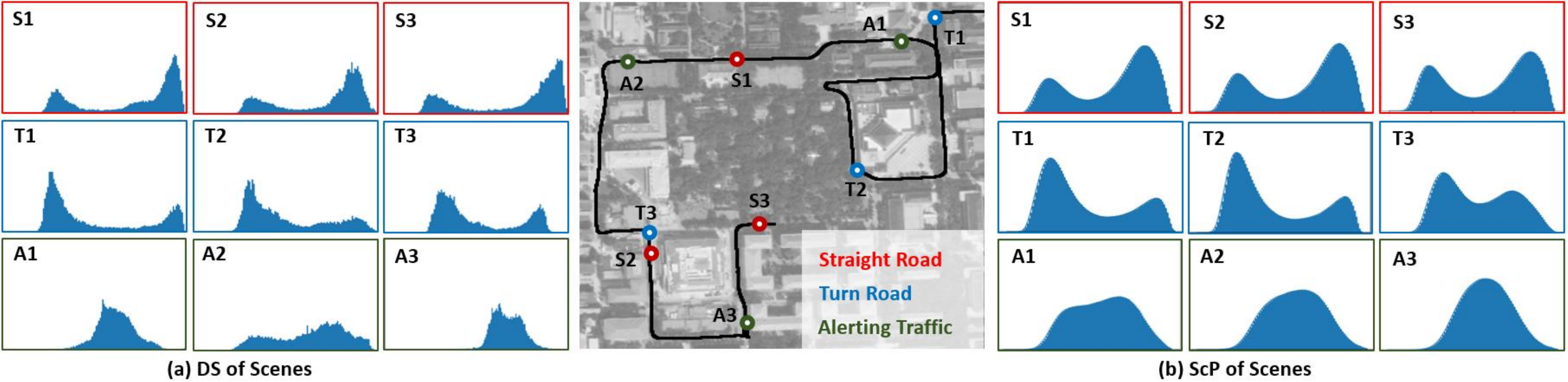}
	\caption{Case study of {\it Distribution of Semantic Scene Similarity} (DS) and {\it Scene Category Profile} (S$c$P).}
	\vspace{-6mm}
	\label{fig:scp}
\end{figure*}

\subsection{Outline of the Method}\label{section:outline}
The method is composed of two modules: 1) semantic scene similarity learning, 2) driving scene profiling and labeling. The workflow is shown in Fig. \ref{fig:meth}.

Semantic scene similarity learning is to find a measure to evaluate the semantic similarity of any two scenes, which is supervised by the anchor points and the {\it neighborhood consistent assumption} for semantically different and similar scenes.
Let $z_i=f_{\theta}(i)$ be a feature encoder converting a high-dimensional image $i$ to a low-dimensional normalized feature vector $z\in \mathbb{Z}^D$. The semantic scene similarity of a pair of image $(i,j)$ is measured in feature space $\mathbb{Z}^D$ on cosine similarity $sim(i,j) = {z_i^T\cdot z_j}$. In feature space $\mathbb{Z}^D$, $z_i$ and $z_j$ are pushed away from each other if they are of two successive anchor points respectively, whereas pulled close if they are in the neighborhood of a same anchor point, and this pull-push process is implemented by contrastive learning. The implementation details are shown in Sec. \ref{section:sssl}.

However, the scenes of a complex driving environment are not clearly separable in feature space $\mathbb{Z}^D$ as shown in Fig. \ref{fig:pca}(b), and the clusters of $z$ might not be semantically meaningful.
This research proposes a method to use the {\it distribution of semantic scene similarity} (DS) as a signature for scene profiling and labeling.
Given a manually selected image frame $q_c$ which presents the typical scene of category $c$, an average DS is estimated as a meta pattern of the scene category (scene category profile, S$c$P). During inference, given an image frame, its DS is first estimated, and then is compared with the S$c$P of each category. The scene label is predicted as the best-match one on the semantic scene similarity distribution. The implementation details are shown in Sec. \ref{section:dspl}.

\subsection{Semantic Scene Similarity Learning}\label{section:sssl}

\subsubsection{Sampling Strategy}For a given query point $q \in [k_{i}, k_{i}+\delta]$ of video $V$, where $k_i$ is an anchor point and $\delta$ is a small integer as mentioned at the beginning of Sec. \ref{section:problem}, its one positive sample and $n$ negative samples are selected by the {\it neighborhood consistent assumption} for contrastive learning training. 

The positive sample $q^+$ is randomly selected near the anchor point $k_i$, that is $q^+ \in [k_{i},k_{i}+\delta]$ . The $n$ negative samples $\{q_j^-|j=1,...,n\}$ are randomly selected near the two successive anchor points of $k_i$, that is $q_j^- \in [k_{i-1},k_{i-1}+\delta] \cup [k_{i+1}, k_{i+1}+\delta]$, as shown in Fig. \ref{fig:meth}.

\subsubsection{Network Design and Loss Function}
Feature encoder $f_\theta$ is implemented by AlexNet\cite{krizhevsky2012imagenet} to convert 3-channel RGB image into 128-dimension feature vector $z\in \mathbb{Z}^{D=128}$, and it is trained by contrastive learning method with InfoNCE loss\cite{oord2018representation}:
\begin{equation}
L = -log\frac{\exp(z_q^T\cdot z_{q^+}/\tau)}{\exp(z_q^T\cdot z_{q^+}/\tau) + \sum_{j=1}^{n}\exp(z_q^T\cdot z_{q^-_j}/\tau)}
\label{equa:loss}
\end{equation}
\subsubsection{Learning Result}As mentioned in Sec. \ref{section:outline}, the semantic scene similarity of a pair of image $(i,j)$ is measured in feature space $\mathbb{Z}^D$ by cosine similarity $sim(i,j) = {z_i^T\cdot z_j}$.

\subsection{Driving Scene Categorization}\label{section:dspl}

\subsubsection{Distribution of Semantic Scene Similarity (DS)}\label{section:d3s}
 \begin{algorithm}[b]
 	\caption{Calculating S$c$P of category $c$}
 	\label{algo:scp}
 	\LinesNumbered
 	\KwIn{$\mathscr{Ds}=\{ds(p)|p \in V\}$, meaning DS of all training video frames}
 	\KwIn{A manually selected typical image $q_c$ of category $c$}
 	\KwOut{S$c$P of category $c$, denoted as $scp(c)$}
 	DSList = []\; 
 	\For{$p \in V$}{
 		\If{{JS-D}$(ds(q_c),ds(p))<\sigma$}{
 			DSList.append($ds(p)$)
 		}
 	}
 	$scp(c)$ = Average(DSList)
 \end{algorithm}

Given a query image $q$, semantic scene similarity is measured between $q$ and each image frame $p$ of the training video $V$, and a set of similarity values are obtained $\Omega(q) = \{sim(q,p) | p \in V\}$. A histogram is subsequently generated on $\Omega(q)$ to find the distribution of semantic scene similarity (DS), which is used as a descriptor of the scene $q$ denoted as $ds(q)$. Such a descriptor captures semantically meaningful pattern as examined in Fig. \ref{fig:meth}.

As shown in Fig. \ref{fig:scp}(a), by randomly choosing semantically typical scenes on straight road, turn road and alerting traffic respectively in different regions, samples of DS are estimated, and it can be found that the DS of the same scene category show similar pattern. The peaks and valleys of the distributions are in fact related with road geometry and dynamic traffic of the scene, which is further investigated through experiments in Sec. \ref{section:expSemanticSceneProfiling}.

Therefore, a method of driving scene profiling and labeling is designed on DS, which is composed of learning semantic scene category profiles, and inferring scene labels for both online and offline applications.

\subsubsection{Scene Category Profile (S$c$P)}\label{section:scp}

A set of DS is estimated $\mathscr{Ds}=\{ds(p)|p \in V\}$ on each image frame $p$ of the training video $V$.
For a semantic scene category $c$, given a typical image frame $q_c$ of it by human operator, the DS of $q_c$, denoted as $ds(q_c)$, is compared with all the $ds(p)$ in $\mathscr{Ds}$ by Jensen-Shannon Divergence $\emph{JS-D}(ds(q_c),ds(p))$ to evaluate the divergence of semantic similarity distributions. The $ds(p)$ is gathered if $\emph{JS-D}(ds(q_c),ds(p))<\sigma$, and an average distribution of the gathered $ds(p)$ are taken as the scene category profile of category $c$, which is denoted as $scp(c)$.
Here $\sigma$ is a pre-define threshold. A pseudo-code is given in Algorithm\ref{algo:scp}

Here is a question: How the selection of the typical image frame $q_c$ of the category $c$ influence $scp(c)$? As shown in Fig. \ref{fig:scp}(b), for each category, several typical scene frames $q_c$ of it are selected to estimate its S$c$P respectively. It can be found that even the typical scene frames $q_c$ of a category are selected at very different locations, the generated S$c$P are very similar. It demonstrates that the method is robust to typical scene frame selection and able to capture the meta pattern of the scene category.

\subsubsection{Scene Category Reasoning}

In reasoning, given the current image frame $q$, $ds(q)$ is estimated by measuring semantic similarity between $q$ and the image frames of the training video $V$ and finding the distribution. Comparing $ds(q)$ with each $scp(c)$ of category $c \in \mathbb{C}$ by Jensen-Shannon Divergence $\emph{JS-D}(ds(q),scp(c))$, the $c$ of the minimal divergence, i.e. the most matched distribution, is found as the predicted label $c_q$ of the current scene $q$, i.e.
\begin{equation}
	c_q= \argmin\limits_{c \in \mathbb{C}} \emph{JS-D}(ds(q),scp(c))
\end{equation}

\section{Experimental Results}\label{section:experimental}

\subsection{Experimental Data and Design}

\begin{figure*}[]
	\centering
	\includegraphics[keepaspectratio=true,width=\linewidth]{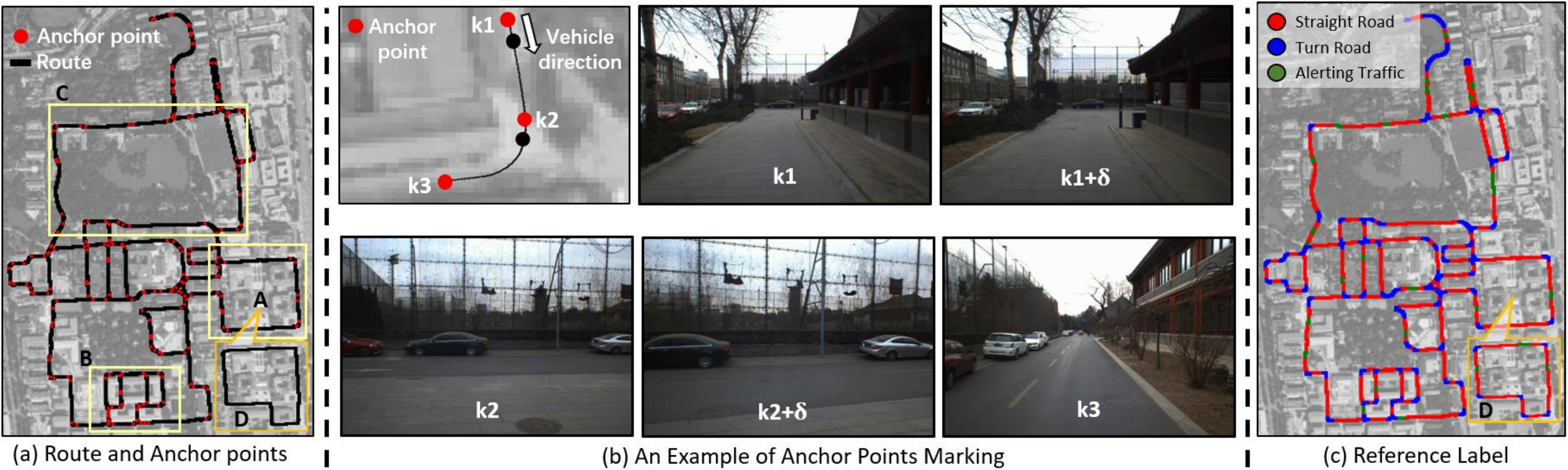}
	\caption{Experimental data illustration.}
	\vspace{-6mm}
	\label{fig:dataset}
\end{figure*}

\subsubsection{Video Data}
An egocentric video is collected in the campus of Peking University by using a front-view monocular RGB camera. Fig. \ref{fig:dataset}(a) shows the driving route that covers a broad and diverse area containing teaching and research zone (A), residential zone (B) and landscape zone (C). Fig. \ref{fig:alertingTraffic} shows typical road scenes, where pedestrians, cyclists and cars are populated, and the behaviors of road users are not strongly confined by traffic rules. 
A 20-minute video is recorded with 36924 image frames at a frame rate of 30 fps, where the teaching and research zone (A) are recorded twice but at different time as it is the start and end position of the driving route. In order to discriminate, we call the second drive at the teaching and research zone as (D).

\subsubsection{Anchor Point}
The video is divided into two parts in experiments as shown in Fig. \ref{fig:dataset}(a).
Part I covers zone A-B-C that is marked by a human driver with anchor points, while part II covers zone D that is unmarked, representing scenes of the same location with zone A but has different routes and dynamic traffics.
A total of 242 anchor points are marked on the video frames by a human driver to imitate his decision making during driving. A new anchor point is marked when the driver recognize a change of the scene attribute, which may cause a different handling in driving or a {\it mental state} change.
Fig. \ref{fig:dataset}(b) shows an example, where the first anchor point (red) is marked at the frame $k_1$ of a straight road scene, the second is at the frame $k_2$ when the vehicle enters a turn, and the third is at the frame $k_3$ after the turn finished. Similarly, anchor points are marked when the driver is alerted by other road users in the scene.
A neighbor frame (black) of $k_1$ and $k_2$ is also shown and denoted as $k_1+\delta$ and $k_2+\delta$ respectively, which demonstrates the consistence of semantic scene attribute in neighborhood.

\subsubsection{Scene Category and Reference Label}\label{section:referenceValue}
In this experiment, scenes are categorized with three semantic labels, i.e. SR: straight road, TR: turn road, AT: alerting traffic. Both SR and AT happen on straight roads, which are differentiated on whether there are other road users who need to be alert in the scene. As the period of the vehicle turning is short, and drivers are usually very concentrated, TR is not further divided into alerting or non-alerting traffics.

Manually assigning semantic labels to each image frame is nontrivial, because the operator will always ask the question: when are the start and the end image frames of the same semantic attribute? Whereas the answer is not clearly defined.
So in this research, we \textbf{do not} have Ground Truth, but developing reference labels via anchor points to assist in analyzing the experimental results.
Since the anchor points are marked whenever a human driver notice a change of scene attribute, although there might be some delay due to the response time of conscious, the anchor points could serve as inaccurate delimiters to divide the stream of image frames into semantically different segments, and each segment is assigned a single label by the human driver while marking the anchor points.
As shown in Fig. \ref{fig:dataset}(c), the reference labels of SR, TR and AT are developed on the image frames of both parts videos.

\subsubsection{Experiment Design}
Experiments are conducted on two levels: (1) Semantic Scene Similarity Learning (Sec. \ref{section:expSssl}). Two experiments are conducted to examine the performance of semantic scene similarity learning, where the scenes are categorized with two and three labels respectively. The results are analyzed at the levels of feature vector, trajectory point and image frame.
(2) Driving Scene Categorization (Sec. \ref{section:drivingSceneCategorization}). The results of semantic scene profiling based on the distribution of similarity evaluation are first explanatory analyzed, and two experiments on scene categorization are then conducted, which aim at demonstrating the performance towards the offline and online applications such as semantic mapping and scene label prediction.

\subsection{Semantic Scene Similarity Learning}\label{section:expSssl}

\begin{figure*}[]
	\centering
	\includegraphics[keepaspectratio=true,width=\linewidth]{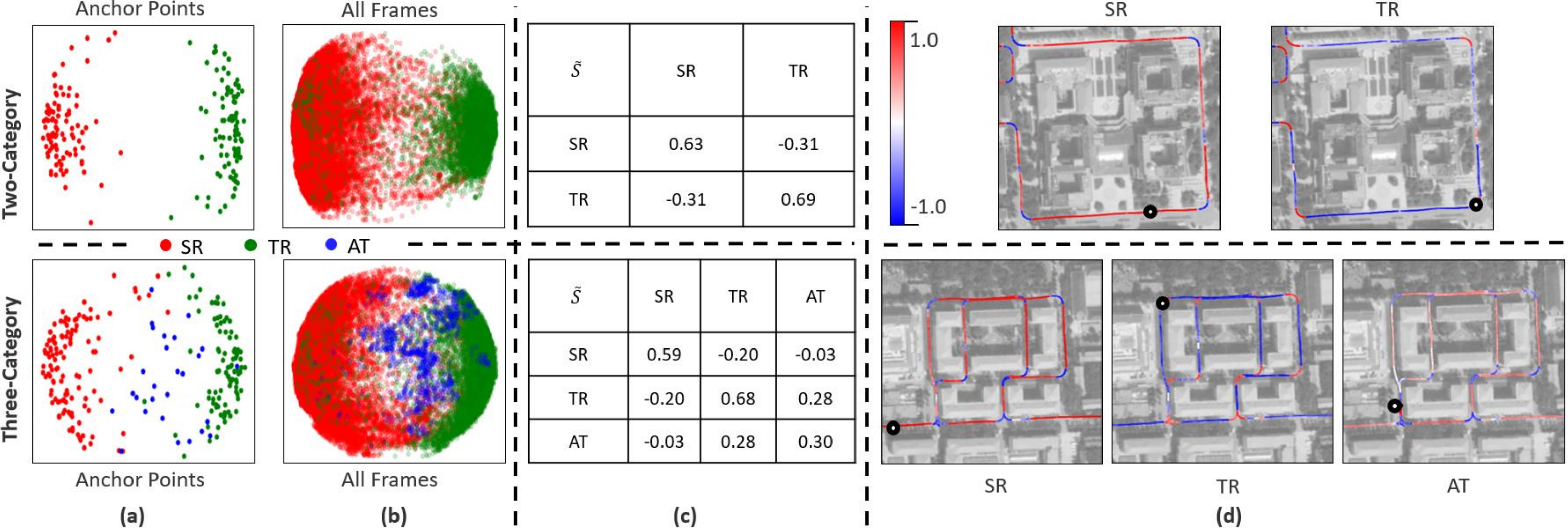}
	\caption{Illustration of semantic scene similarity at the level of feature and trajectory. (a)\&(b) Visualization of feature  after dimension reduction. (c) Similarity of scene categories. (d) Similarity between the query frame (black point) and trajectory.}
	\label{fig:pca}
	\vspace{-4mm}
\end{figure*}

\subsubsection{Metrics Learning}

In order to examine the performance of semantic scene similarity learning, a two-category experiment is specially conducted, in addition to the three-category one as designed in the previous subsection.
In this experiment, the two scene categories are straight road (ST) and turn road (TR). The anchor points alerting traffic (AT) are removed from the original set, and the reference labels AT of the video segments are changed to ST.
The results of both two- and three-category learning are shown in Fig. \ref{fig:pca}.

Given the video part I and the set of marked anchor points, each experiment learns a feature representation $f_{\theta}$ following the method described in Sec. \ref{section:sssl}, where for each query point $q$, one positive and $n=16$ negative samples are randomly taken to estimate the InfoNCE loss of Eqn.\ref{equa:loss}.
With the learned $f_{\theta}$, each image is converted into a feature vector $z \in \mathbb{Z}^D$, where $D=128$ in this research.
For visualization, these feature vectors are further reduced to 2 dimension using PCA, and colored on their reference labels and shown in Fig. \ref{fig:pca}(a)\&(b). 
In Fig. \ref{fig:pca}(a), it can be found that the anchor points are clustered in different zones that are consistent with their reference labels (i.e. semantic meaning) in both two- and three-category results.
Whereas when testing all image frames, as visualized in Fig. \ref{fig:pca}(b), the feature vectors of different colors can not be easily separated.

\subsubsection{Performance Analysis}
To further analyze the results, a new metric is defined as below to evaluate similarity of two scene categories:
$$\tilde{S}(c_1, c_2)=\sum_{p\in c_1, q \in c_2} \frac{1}{|c_1|\cdot |c_2|}\cdot sim(p,q)$$
where $\tilde{S}(c_1,c_2) \in [-1,1]$, $|c_1|$ and $|c_2|$ denote the image number of category $c_1$ and $c_1$ respectively, and $sim(p,q)$ is the semantic scene similarity of two images on cosine similarity of their feature vectors $z_p$ and $z_q$ as mentioned before. The lower the $\tilde{S}$, the more separable of the two categories and the better performance of the learned $f_{\theta}$.
As shown in Fig. \ref{fig:pca}(c), comparing to the diagonal values representing intra-category similarities, the off-diagonal ones of inter-category similarities are much lower, meaning that the learning method has efficiency from a statistical viewpoint.

On the other hand, as shown in Fig. \ref{fig:pca}(d), randomly select a query frame $q$ (black point) with reference label of SR, TR or AT, find its similarity with all other image frames of the video, i.e. $\forall p \in V, sim(p,q)$, and visualize the similarity values by coloring the trajectory points. Red for similar (1.0), blue for dissimilar (-1.0). The colors change with the selected query frame $q$.
Given a query frame of reference label SR, it can be found that all trajectory points on turn roads are blue, and the straight roads are mainly red. Given a query frame of reference label TR, the trajectory points are mainly red on turn roads whereas blue on straight roads. Given a query frame of reference label AT, most trajectory points are colored from blue to light red.

\begin{figure}[h]
	\centering
	\includegraphics[keepaspectratio=true,width=\linewidth]{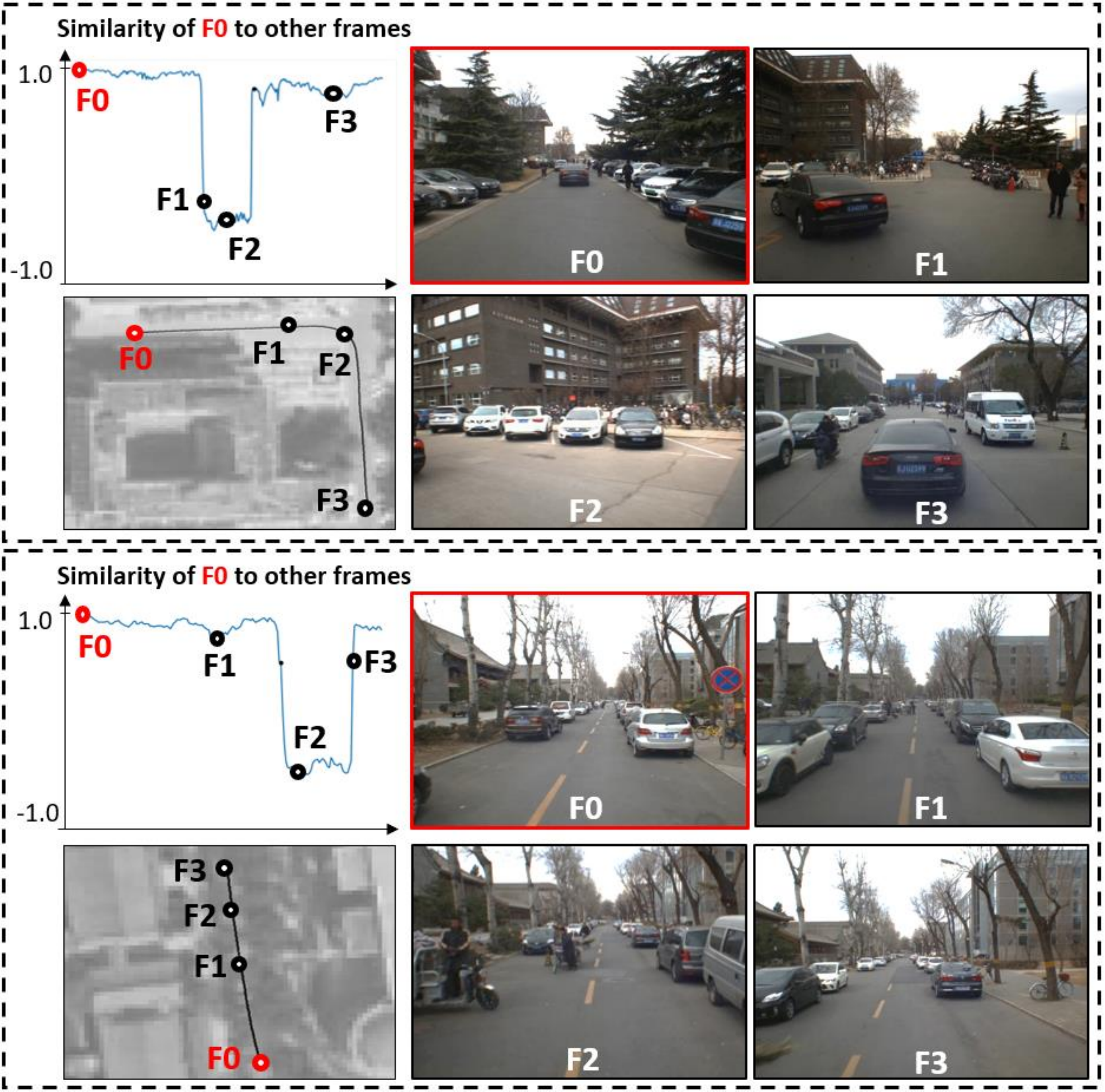}
	\caption{Illustration of semantic scene similarity at the level of image frame.}
	\label{fig:continues}
\end{figure}

Fig. \ref{fig:continues} further illustrates the similarity values at the level of image frames. Two cases of traversing through a turn road (top of Fig. \ref{fig:continues}) and alerting traffic (bottom of Fig. \ref{fig:continues}) are shown respectively, where F0 is the query frame $q$ of normal straight road scenes (SR). The similarity values of the subsequent frames with F0 are plotted, which show a dramatic down-and-up pattern, and we can find that the image frames of lower similarity values are scenes of semantically different attributes with F0, e.g. F1-F2 are turn road in the top case and F2 is alerting traffic in the bottom case.

The above results show that the proposed method has efficiency in learning a semantically meaningful metric to evaluate the similarity/distance of dynamic driving scenes. 

\subsection{Driving Scene Categorization}\label{section:drivingSceneCategorization}
	
\subsubsection{Semantic Scene Profiling}\label{section:expSemanticSceneProfiling}

\begin{figure}[t]
	\centering

	\includegraphics[keepaspectratio=true,width=\linewidth]{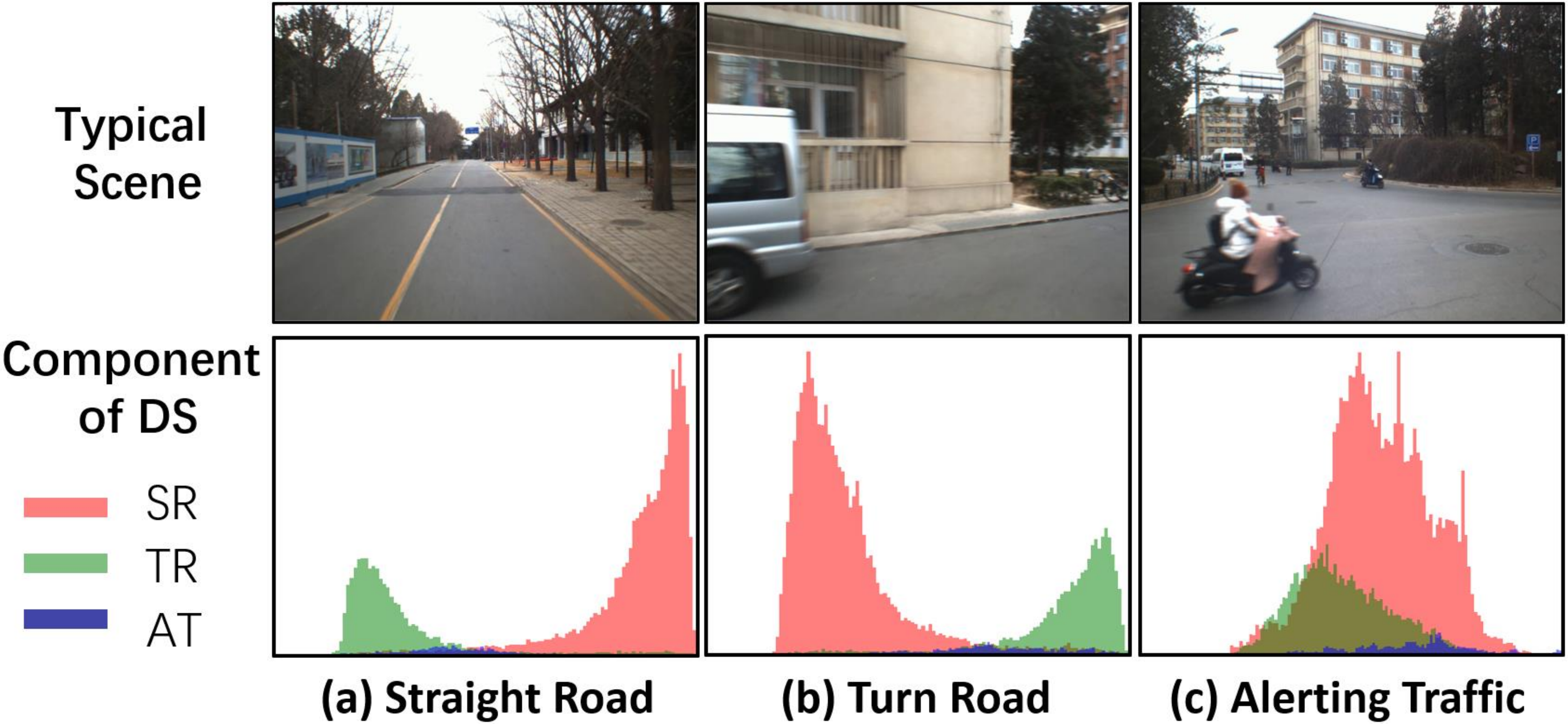}
	\caption{Explanatory analysis of semantic scene profiling based on the distribution of similarity.}
	\vspace{-4mm}
	\label{fig:ds3}
\end{figure}

Based on the distribution of semantic scene similarity (DS), profiles are generated for scene categories (S$c$P) of SR, TR and AT via typical image frames as shown in Fig. \ref{fig:scp}.
The peaks and valleys of the DS and S$c$P are related with the road geometry and dynamic traffic of the scenes.
Fig. \ref{fig:ds3} examines the components of the DS of each category, where the DS is estimated by first selecting a typical scene $q$, and then estimating the semantic scene similarity $\Omega(q) = \{sim(q,p) | p \in V\}$, and finally generating a histogram on $\Omega(q)$.
The similarity values can be divided into three types according to their reference labels, therefore colored bar charts can be generated to visualize the contributions of each component.

Take straight road scenes as example shown in Fig. \ref{fig:ds3}(a), the right (high similarity) and left (low similarity) peaks of DS are respectively contributed from straight and turn road scenes, and the area of the peaks reflects the proportion of straight and turn roads in the scene frames, e.g. in an environment dominated by straight roads, the right peak could be very sharp and tall, whereas in an environment of various road types, the right peak could be wide and flat. Turn road scenes are the opposite shown in Fig. \ref{fig:ds3}(b). The DS of alerting traffic is different from straight and turn roads as shown in Fig. \ref{fig:ds3}(c), which is a single-peak pattern.
Comparing to straight and turn road scenes, alerting traffic is a minority, whereas the density and distribution could be a significant feature to describe the dynamic traffic of the scene.

\begin{figure}[b]
	\centering
	\vspace{-4mm}
	\includegraphics[keepaspectratio=true,width=\linewidth]{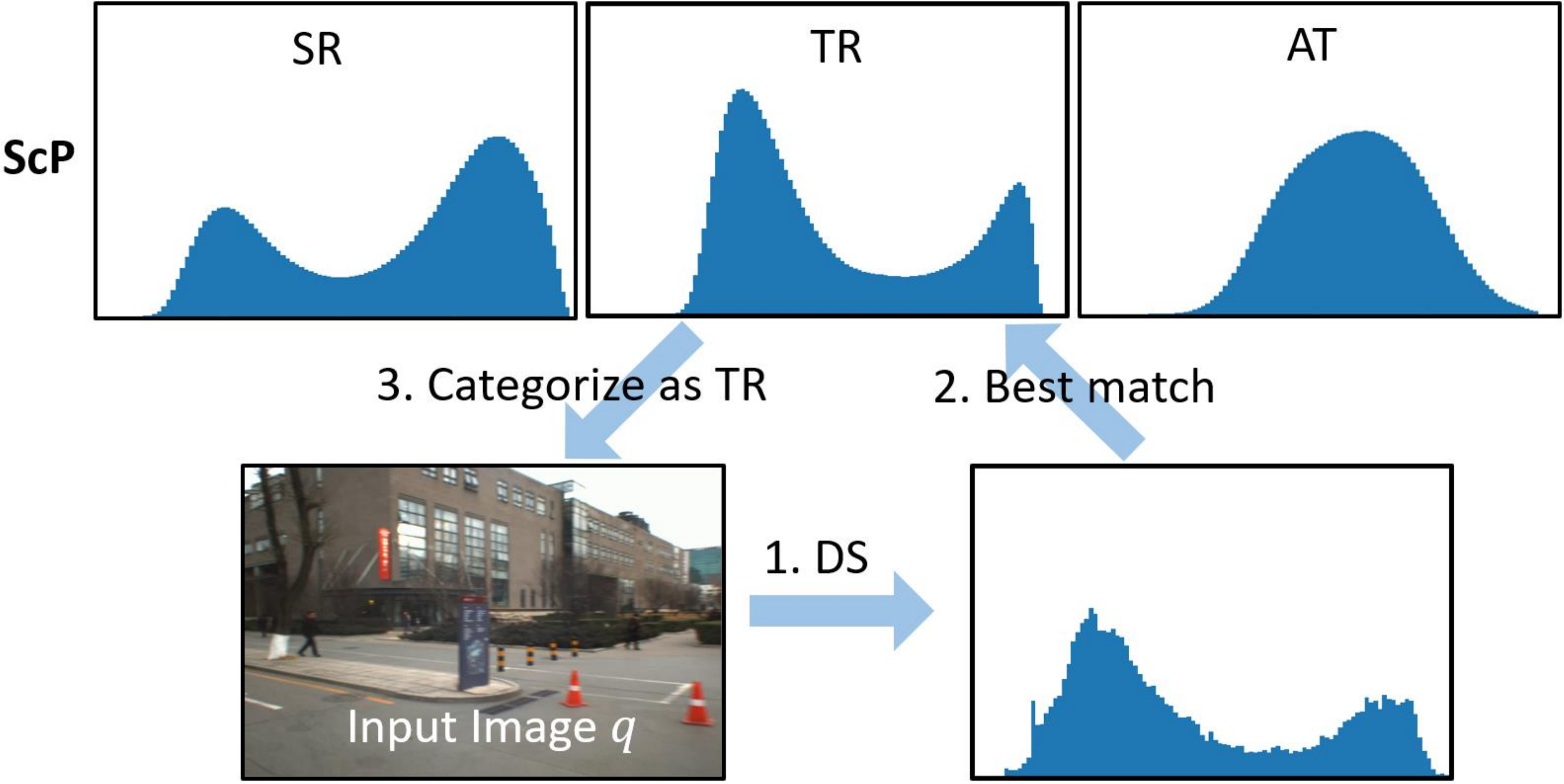}
	\caption{Categorization procedure.}
	\label{fig:classifyCase}
\end{figure}

\subsubsection{Scene Categorization}

Experiments are conducted on video part I with marked anchor points for learning and video part II of new scenes, aiming at performance validation towards offline applications such as semantic mapping and online applications such as scene label prediction. 
Both experiments are conducted in the same procedure that is demonstrated in Fig. \ref{fig:classifyCase} by case result.
Given an image frame $q$ of an offline video or from an online camera, its DS is first estimated by evaluating semantic similarity with the image frames of the training video, and then compared with the scene category profiles (S$c$P) to find the category $c$ of the best match $scp(c)$. 
For a given $224\times 224\times3$ RGB image, it take 50 ms to get its feature on NVIDIA GTX 1080, and 0.8 ms to calculate its DS, and 0.2 ms to find the best match S$c$P of all categories on Intel Core i9 using Python.

\begin{figure}[b]
	\centering
	\vspace{-2mm}
	\includegraphics[width=0.5\textwidth]{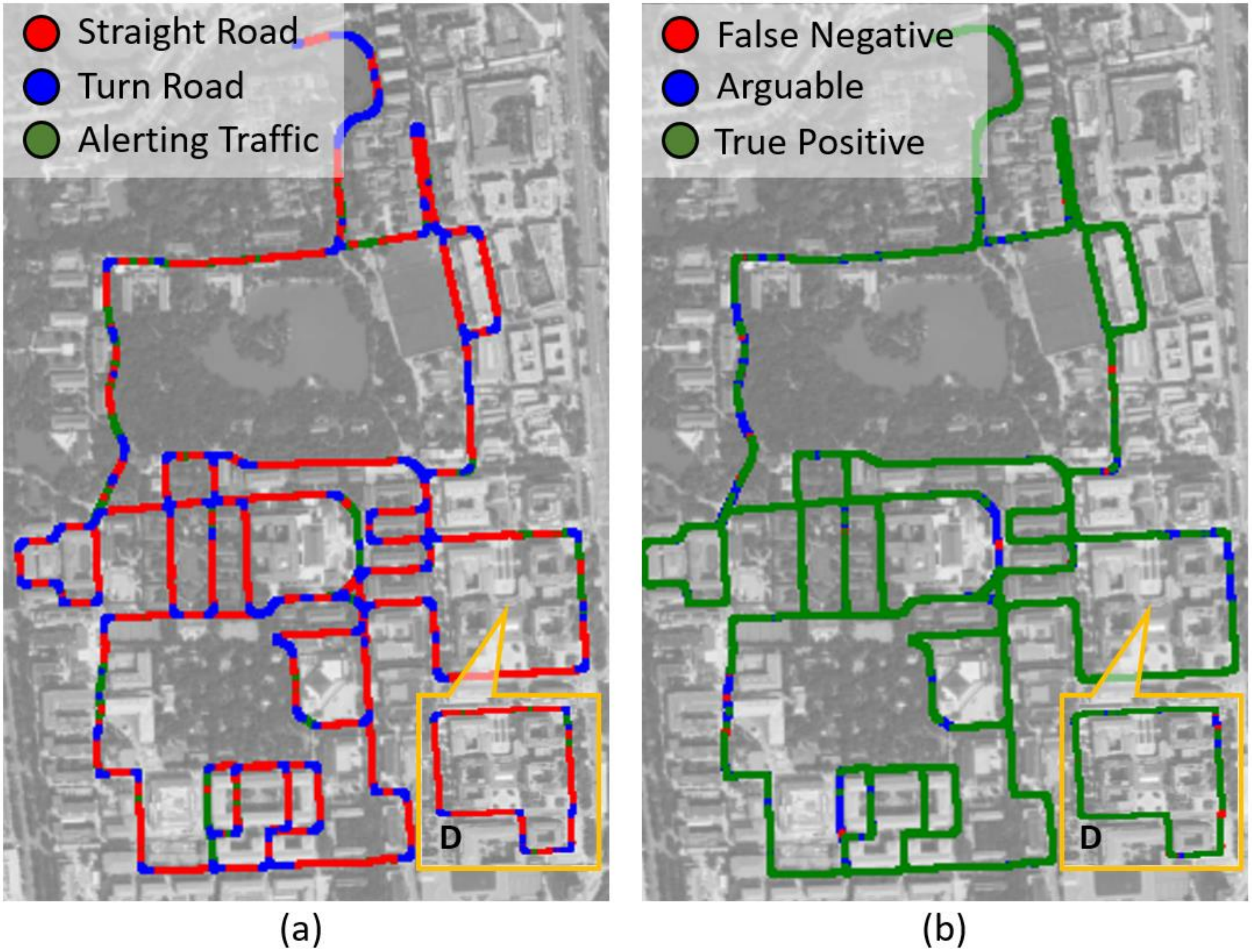}
	\caption{Scene categorization results. (a) The predicted label. (b) Comparing with reference label.}
	\label{fig:classificationTraj}
\end{figure}

\begin{figure}[t]
	\centering
	\setlength{\belowcaptionskip}{-4mm}
	\includegraphics[keepaspectratio=true,width=\linewidth]{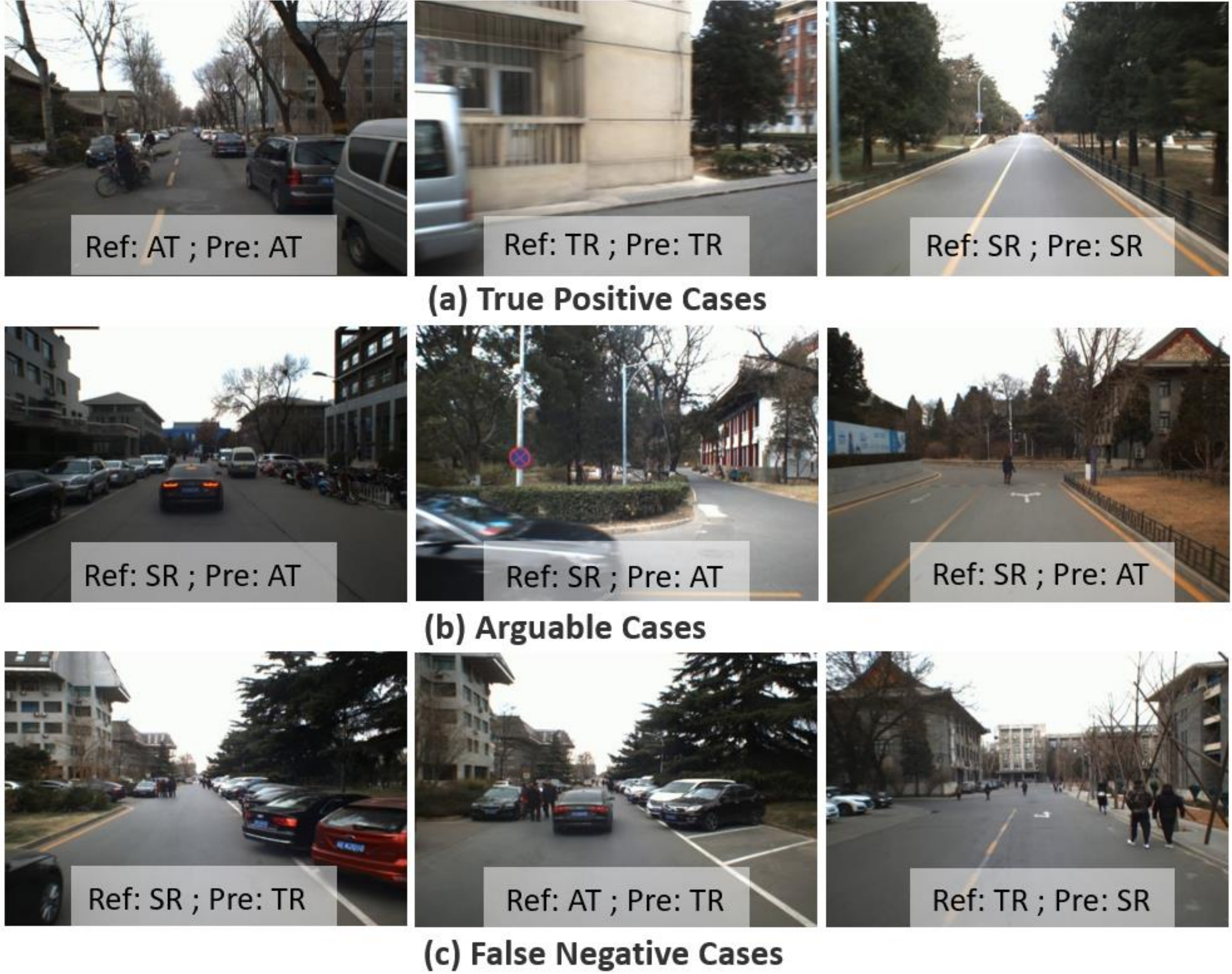}
	\caption{Categorization results illustration.}
	\label{fig:failCase}
\end{figure}

Scene categorization results on both video part I and II are shown in Fig. \ref{fig:classificationTraj}(a). By comparing with the reference labels, the results are divided into three types as shown in Fig. \ref{fig:classificationTraj}(b). \textbf{True Positive (TP)} are those matched with the reference labels. The rest set of unmatched results is examined by a human driver, where two different types are discovered: \textbf{False Negative (FN)} are those of wrong categorizations, while \textbf{Arguable (AG)} are arguable results.
Fig. \ref{fig:failCase} shows examples of \textbf{True Positive (TP)}, \textbf{False Negative (FN)} and \textbf{Arguable(AG)} results where Ref is the reference label and Pre is the predicted label.

Since reference labels are generated on the set of anchor points marked when watching the video, the human decision is made on \emph{image sequences}, so even a driver just encounters a situation like Fig. \ref{fig:failCase}(b), the scenes will not be marked as alerting traffic, as the decision is made based on the prediction of other road users' tendency. However, this research develops a \emph{single-image-based} algorithm.  Therefore, such results are counted as \textbf{Arguable (AG)} as they are reasonable results with the input of single images.

\begin{table}[h]
	\centering
	\caption{Three types of categorization results.}
	\begin{tabular}{|c|c|c|c|c|}
		\hline
		& Anchor points & TP       & AG       & FN       \\ \hline
		Video part I & \cmark & 74.82\% & 22.35\% & 2.83\%  \\ \hline
		Video part II & \xmark & 56.64\% & 28.80\% & 14.56\% \\ \hline
	\end{tabular}
	
	\label{tab:class}
\end{table}

These three types of results are counted in TABLE \ref{tab:class}. The video part I is marked with anchor points for learning, and this experiment demonstrates the performance for offline applications, and the percentage of \textbf{TP}, \textbf{AG} and \textbf{FN} are 74.82\%, 22.35\% and 2.83\% respectively. By calculating \textbf{AG} with \textbf{TP}, the positive results are 97.17\%. In the experiment on video part II that of new scenes, the positive results are 56.64\%(TP) + 28.80\%(AG)=85.44\%.

\section{CONCLUSIONS}\label{section:conclusions}
	This paper proposes a method of task-driven driving scene categorization using weakly supervised data.
	Given a front-view video of a driving scene, a set of anchor points is marked by following the decision making of a human driver, where an anchor point is not a semantic label but an indicator meaning the semantic attribute of the scene is different from that of the previous one.
	A measure is learned to discriminate the scenes of different semantic attributes via contrastive learning, and a driving scene profiling and categorization method is developed by modeling the distribution of semantic scene similarity based on that measure.
	The proposed method is examined by using a front-view monocular RGB video that is recorded when a vehicle traversed the cluttered dynamic campus of Peking University.
	The video is separated into two parts, where Part I is marked with anchor points for learning and examining the performance of scene categorization towards offline applications, and Part II contains new scenes and is used to examine the performance for online prediction. Scenes are categorized into straight road, turn road and alerting traffic that are demonstrated with example images. The results of semantic scene similarity learning and driving scene categorization are extensively studied, and positive results of 97.17 \% of scene categorization on the video in learning and 85.44\% on the video of new scenes are obtained.
	Future work will be addressed to extend the method to use a video clip or a stream of history images as the input, which provides temporal cues to model and reason the dynamic change of the scenes and realize a more human-like scene categorization.

\bibliographystyle{IEEEtran}
\bibliography{ref}
\end{document}